\documentclass{article} 
\usepackage{iclr2017_conference,times}
\usepackage{amsfonts}
\usepackage{amsmath}
\usepackage{hyperref}
\usepackage{url}
\usepackage{amsmath}
\usepackage{graphicx}

\usepackage[utf8]{inputenc}

\DeclareFixedFont{\ttb}{T1}{txtt}{bx}{n}{12} 
\DeclareFixedFont{\ttm}{T1}{txtt}{m}{n}{12}  

\usepackage{color}
\definecolor{deepblue}{rgb}{0,0,0.5}
\definecolor{deepred}{rgb}{0.6,0,0}
\definecolor{deepgreen}{rgb}{0,0.5,0}

\usepackage{listings}

\newcommand\pythonstyle{\lstset{
language=Python,
basicstyle=\ttm,
otherkeywords={self},             
keywordstyle=\ttb\color{deepblue},
emph={MyClass,__init__},          
emphstyle=\ttb\color{deepred},    
stringstyle=\color{deepgreen},
frame=tb,                         
showstringspaces=false            %
}}

\lstnewenvironment{python}[1][]
{
\pythonstyle
\lstset{#1}
}
{}

\newcommand\pythoninline[1]{{\pythonstyle\lstinline!#1!}}

\newcommand{\norm}[1]{\left\lVert#1\right\rVert}

\title{On reproduction of \\ \emph{On the regularization of Wasserstein GANs}}

\author{Junghoon Seo \\
R\&D Center, Satrec Initiative\\
Daejeon, 34051, Republic of Korea \\
\texttt{sjh@satreci.com}
\And Taegyun Jeon \\
R\&D Center, Satrec Initiative\\
Daejeon, 34051, Republic of Korea \\
\texttt{tgjeon@satreci.com}
}

\begin{document}

\maketitle

\begin{abstract}
This report has several purposes. First, our report is written to investigate the reproducibility of the submitted paper \emph{On the regularization of Wasserstein GANs} (2018). Second, among the experiments performed in the submitted paper, five aspects were emphasized and reproduced: learning speed, stability, robustness against hyperparameter, estimating the Wasserstein distance, and various sampling method. Finally, we identify which parts of the contribution can be reproduced, and at what cost in terms of resources. All source code for reproduction is open to the public. 
\end{abstract}

\section{Introduction}

This report is a submission for participating \emph{ICLR 2018 Reproducibility Challenge}.\footnote{For more information about the challenge, please visit the following website: \url{http://www.cs.mcgill.ca/~jpineau/ICLR2018-ReproducibilityChallenge.html}.} The target paper of reproducibility verification is \cite{anonymous2018on}. The target paper proposes a new regularization term that stabilizes training of generative adversarial network (GANs) based on 1-Wasserstein metric and makes it robust to selection of hyperparameters.

\subsection{Wasserstein distance and regularization}

Generative adversarial networks, which is dealt as one of the most attractive frameworks among generative models, is first introduced in \cite{NIPS2014_5423}. It is well known that training process of GANs could be interpreted as metric minimization between data sample and generated sample. For example, the original GANs paper argues that its training process is equivalent to minimization of Jensen-Shanon distance between data sample and generated sample. Moreover, some works of literatures such as \cite{NIPS2016_6066, zhao2016energy, li2017mmd, chen2016infogan} have visited various metric optimization other than the original's Jensen-Shanon distance.

\cite{arjovsky2017wasserstein} proposed Wasserstein GAN (WGAN), which is that the concept of Wasserstein distance is introduced into GANs framework. The authors argued that optimizing Wasserstein distance has advantages over other metrics that measure the difference between the probability measures. In \cite{arjovsky2017wasserstein}, this advantage is described well in Theorem 1 and theoretically justified in Appendix C. However, some uncapacitated minimum cost flow algorithms like \cite{orlin2013max}, which are commonly used computing optimal transport distance, are highly intractable. According to \cite{villani2008optimal}, infimum problem for computing Wasserstein distance could be converted as supremum problem thanks to Kantorovich-Rubinstein duality. The derived optimization problem for adversarially solving optimal transport problem is the following:
\begin{equation} \label{eq:1}
\min_{\mathbb{P}_\theta}{\max_{f_\phi \in Lip_{1}}{\mathbb{E}_{x\sim\mathbb{P}_r}[f_\phi(x)]-\mathbb{E}_{y\sim\mathbb{P}_\theta}[f_\phi(y)]}}
\end{equation}
where each notation refers:
\begin{itemize}
\item $\mathbb{P}_r$ is the probability distribution of sample data.
\item $\mathbb{P}_\theta$ is the probability distribution of generated data, which is modeled by the generator neural network and parameterized by $\theta$.
\item $f_\phi$ is the critic function, which is modeled by the discriminator neural network and parameterized by $\phi$.
\item $Lip_{1}$ is the set of all 1-Lipschitz functions.
\end{itemize}

The most critical part of equation (\ref{eq:1}) is that the critic function should have 1-Lipschitz continuity. This is because specifying Lipschitz constant of the continuous function is computationally feasible. In \cite{arjovsky2017wasserstein}, weight clipping technique is used to enforce 1-Lipschitz continuity for the critic function but it is rarely effective. Thus, some follow-up studies face this problem.

\cite{gulrajani2017improved} proposed a new regularization term for enforcing convergence on the gradient of data point, with sampling strategy on interpolation between the sample point and the generated point. \cite{kodali2017on} modified the sampling strategy into local perturbation on sample data. \cite{anonymous2018on} improved the regularization term by relieving convergence of gradient when the gradient of a sampled point is under one.

Taken together, the ultimate optimization problems in each paper are:
\begin{enumerate}
\item \cite{arjovsky2017wasserstein} (\emph{WGAN})
\begin{equation} \label{eq:2}
\min_{\theta}{\max_{\phi \in [-c, +c]}{\mathbb{E}_{x\sim\mathbb{P}_r}[f_\phi(x)]-\mathbb{E}_{y\sim\mathbb{P}_\theta}[f_\phi(y)]}}
\end{equation}
where $c \in \mathbb{R}^+ \ll 1$.
\item \cite{gulrajani2017improved} (\emph{WGAN-GP})
\begin{equation} \label{eq:3}
\min_{\theta}{\max_{\phi}{\mathbb{E}_{x\sim\mathbb{P}_r}[f_\phi(x)]-\mathbb{E}_{y\sim\mathbb{P}_\theta}[f_\phi(y)]} + \lambda \mathbb{E}_{z}[(\norm{\nabla f_\phi(z)}_2 - 1)^2]}
\end{equation}
where $z = \alpha x + (1-\alpha) y$, $x \sim \mathbb{P}_r$, $y \sim \mathbb{P}_\theta$, $\alpha \sim \emph{U}[0, 1]$, and $\lambda \in \mathbb{R}^+$.
\item \cite{kodali2017on} (\emph{DRAGAN})\footnote{The content of recent revision version of this paper differs greatly from one of the original arXiv version. The sampling strategy is also one of the major differences between the two versions. This report follows the old version because we think that the reproduction paper probably followed the old version.}
\begin{equation} \label{eq:4}
\min_{\theta}{\max_{\phi}{\mathbb{E}_{x\sim\mathbb{P}_r}[f_\phi(x)]-\mathbb{E}_{y\sim\mathbb{P}_\theta}[f_\phi(y)]} + \lambda \mathbb{E}_{z = x+\alpha \delta, x \sim \mathbb{P}_r, \alpha \sim \emph{U}[0, 1]}[(\norm{\nabla f_\phi(z)}_2 - 1)^2]}
\end{equation}
where $\delta \sim C \cdot real\_minibatch.std() \cdot \emph{U}[0, 1]$,  and $\lambda, C \in \mathbb{R}^+$.
\item \cite{anonymous2018on} (\emph{WGAN-LP}), which is the reproduction target on this report
\begin{equation} \label{eq:5}
\min_{\theta}{\max_{\phi}{\mathbb{E}_{x\sim\mathbb{P}_r}[f_\phi(x)]-\mathbb{E}_{y\sim\mathbb{P}_\theta}[f_\phi(y)]} + \lambda \mathbb{E}_{z}[(\max(0, \norm{\nabla f_\phi(z)}_2 - 1))^2]}
\end{equation}
where $z$ follows the sampling strategy which is used in either Equation (\ref{eq:3}) or Equation (\ref{eq:4}), and $\lambda \in \mathbb{R}^+$.
\end{enumerate}

\subsection{The target questions and experimental methodology}

The ultimate goal of all experiments in \cite{anonymous2018on} is that Equation (\ref{eq:5}) is better than Equation (\ref{eq:3}) and Equation (\ref{eq:4}) in aspect of training stability and convergence speed to optimize optimal transport problem. The detailed target questions of this reproducibility report are listed up: (All subsection names appears in the section \emph{experiment} of \cite{anonymous2018on}.)
\begin{enumerate}
\item Does the critic learn the function faster?
\begin{itemize}
\item Subsection \emph{Level sets of the critic}.
\end{itemize}
\item Does the critic learn the function more stably?
\begin{itemize}
\item Subsection \emph{Evolution of the critic loss}.
\end{itemize}
\item Is learning of the critic more robust against change of hyperparameter $\lambda$?
\begin{itemize}
\item Subsections \emph{Level sets of the critic} and \emph{Evolution of the critic loss}.
\end{itemize}
\item Is the optimal transport problem solved more well?
\begin{itemize}
\item Subsection \emph{Estimating the Wasserstein distance}.
\end{itemize}
\item Is the learning more robust against change of sampling method?
\begin{itemize}
\item Subsection \emph{Different sampling methods}.
\end{itemize}
\end{enumerate}
Note that subsection \emph{Sample quality on CIFAR-10} in the experiment section of \cite{anonymous2018on} and Appendix D.5 \emph{Optimizing the Wasserstein-2 distance} are out of our reproduction scope.\footnote{At the beginning of the reproduction, we refer to the arXiv uploaded version of the paper. However, we noticed that the revised version of OpenReview recently added subsection \emph{Sample quality on CIFAR-10}. As a matter of time, the experiments in this subsection were excluded from the scope of this report. On the other hand, we are having trouble implementing a regularization loss term that minimizes the Wasserstein-2 distance in \emph{Tensorflow}.
}

Please refer to the experimental section of the original article to see the intent of each experiment.

\section{Details of implementation}
Since the implementation of the target article is not public, we have implemented it firsthand. Before revealing the detailed implementation, we leave some Github repository links that could be helpful for the implementation of the paper. All the repositories are implemented in Python and Tensorflow.
\begin{enumerate}
\item @igul222/\emph{improved\_wgan\_training}\footnote{\url{https://github.com/igul222/improved_wgan_training}} : Implementation of \cite{gulrajani2017improved}.
\item @kodalinaveen3/\emph{DRAGAN}\footnote{\url{https://github.com/kodalinaveen3/DRAGAN}} : Implementation of \cite{kodali2017on}.
\item @hwalsuklee/\emph{tensorflow-generative-model-collections}\footnote{\url{https://github.com/hwalsuklee/tensorflow-generative-model-collections}} : Implementation of GANs variants.
\end{enumerate}

All source code in this report is open to the public on \url{https://github.com/mikigom/WGAN-LP-tensorflow}. We have implemented the experiments on Tensorflow 1.3.0. The experiments are conducted on Intel(R) Xeon(R) CPU E5-1650 v4 @3.60GHz and GeForce GTX 1080 Ti Graphics Cards.

\subsection{Project structure}
The project consists of four Python modules: \emph{data\_generator.py}, \emph{model.py}, \emph{reg\_losses.py}, and \emph{trainer.py}. \emph{data\_generator.py} is a module that provides a class that generates the sample data needed for learning. \emph{model.py} is a module that implements 3-layer neural networks for a generator and a critic. \emph{reg\_losses.py} defines the sampling method and loss term for regularization. \emph{trainer.py} includes a pipeline for model learning and visualization.

\subsection{Notes for our implementation}
\begin{itemize}
\item All hyperparameters followed those presented in the original paper. The hyperparameters of the RMSProp optimizer, the only unknown hyperparameters (except for learning rate), followed the \emph{Tensorflow} default values.
\item \emph{data\_generator.py} includes Python class-form generators into which code of the data generator in @igul222/\emph{improved\_wgan\_training} is refactored. Three sample distributions, \emph{8Gaussians}, \emph{25Gaussians}, and \emph{Swiss Roll}, are implemented. For generating \emph{Swiss Roll} dataset, \emph{sklearn.datasets.make\_swiss\_roll()} is used.
\item \emph{model.py} is implemented in Python class form with TF-slim.
\item All sample perturbation methods in \emph{reg\_losses.py} are implemented in \emph{Tensorflow}, not in \emph{numpy}.
\item For drawing level sets and 2-D data, \emph{matplotlib} is used. For visualizing loss, \emph{Tensorboard} summary operator is simply used.
\item \emph{scipy.optimize.linear\_sum\_assignment()} is used for computing earth mover's distance(EMD). It is equivalent to Hungarian method, or Kuhn–Munkres algorithm.
\end{itemize}

To see our implementation in more detail, please check out the repository above mentioned. 
\section{Analysis and discussion of reproduction}

We first want to specify that the experiment follows the trends presented in the paper as a whole, but the overall learning speed is relatively slow. It is assumed that this is due to differences of hyperparameter in RMSProp, or in unrecognized elements. However, since this is not very inconsistent with the overall tendency of the experiment, we proceeded to reproduce the experiments without searching to solve it. Thus, we reproduced the figures from 500, 2500, 5000, 10000 iteration for drawing figures of the subsection 3.1 \emph{Level sets of the critic}. (instead of the iteration presented in the original paper). In order to reproduce experiments of subsection 3.2 \emph{Estimating the Wasserstein distance}, the learning was done up to 20k steps. However, in the case of reproduction of subsection 3.3 \emph{Estimating the Wasserstein distance}, the training is done up to the 2k step (which is the same value with the original paper), since the calculation of EMD takes a considerable amount of time.

It takes about 12 minutes to learn the 20k steps without EMD calculation, and it takes about 2 hours to learn the 2k steps when EMD calculation is included. In this report, we have simplified some experiments somewhat and decided to skip the part where the median of the critic's negative loss was shown by repeating the experiment and show only the single run results.  Note that the critic's negative loss in the original paper and the implementation in this report is the opposite of the sign.

In the original paper, it is not described in which data set the experiment was carried out in experiments other than those included in subsection \emph{Level sets of the critic}. In this report, \emph{Swiss Roll} dataset is used.

\subsection{Level sets of the critic}

\begin{figure}[h]
\label{fig:1}
\begin{center}
\includegraphics[width=350px]{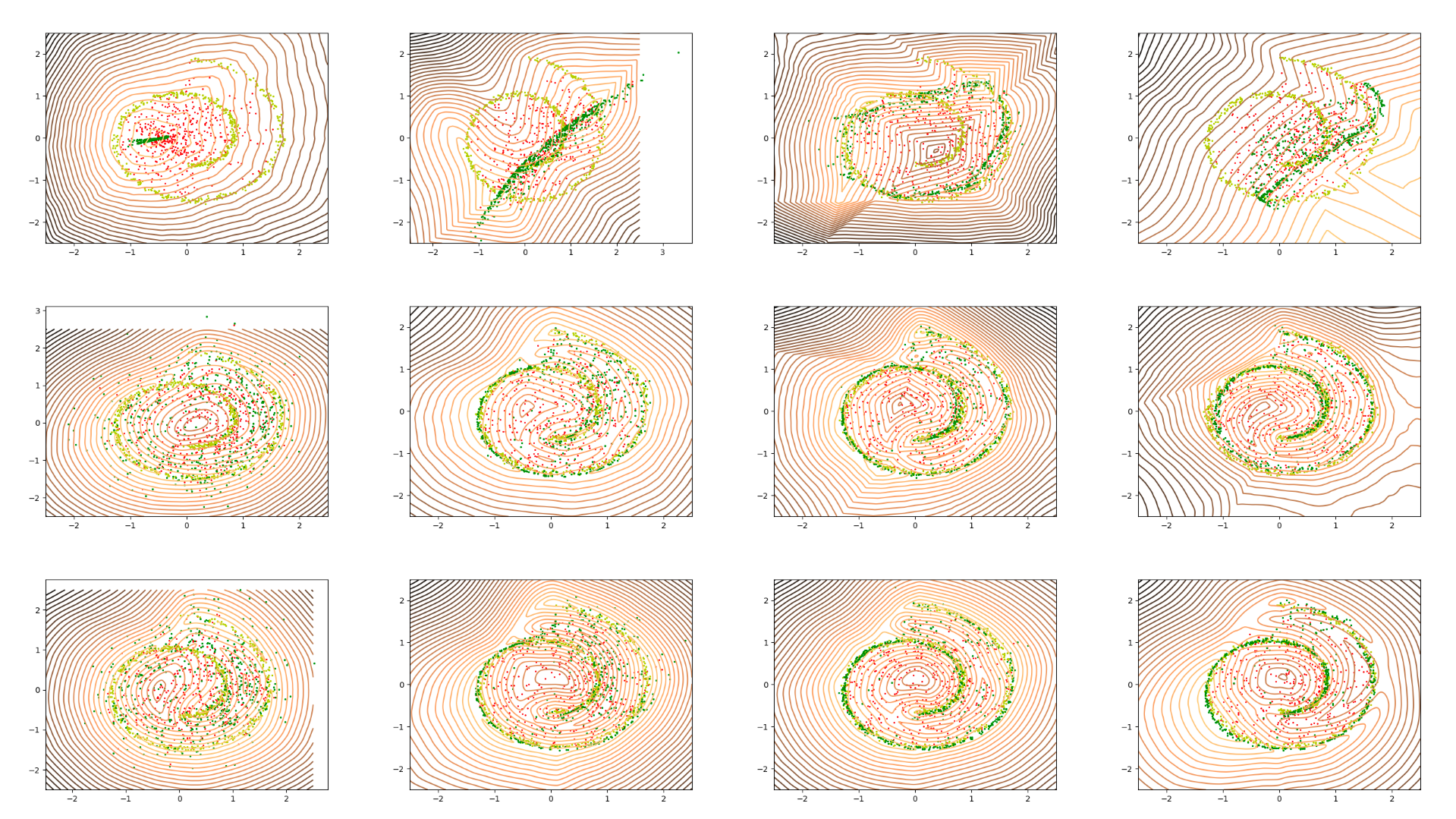}
\end{center}
\caption{Reproduction results of the original paper's Fig. 3. Level sets of the critic f of WGANs during training on \emph{Swiss Roll} dataset. The bright line corresponds to high, dark line to low values of the critic. Training samples are indicated in yellow, generated samples in green, and samples used for the penalty term in red. Top: GP-penalty with $\lambda = 10$. Middle: GP-penalty with $\lambda = 1$. Bottom: LP-penalty with $\lambda = 10$.}
\end{figure}

Fig. 1, Fig. 4, and Fig. 5 correspond to the reproduction results of this section. It can confirm that WGAN-LP is robust to the selection of $\lambda$ than WGAN-GP. Unlike the other two datasets, in WGAN-GP training in \emph{8Gaussians} under $\lambda = 10$, it is consistent with our original thesis and implementation that the critic neural network leans the correct function ultimately. (although of course slower and unstable than other conditions.)

\subsection{Evolution of the critic loss}

\begin{figure}[h]
\label{fig2}
\begin{center}
\includegraphics[width=350px]{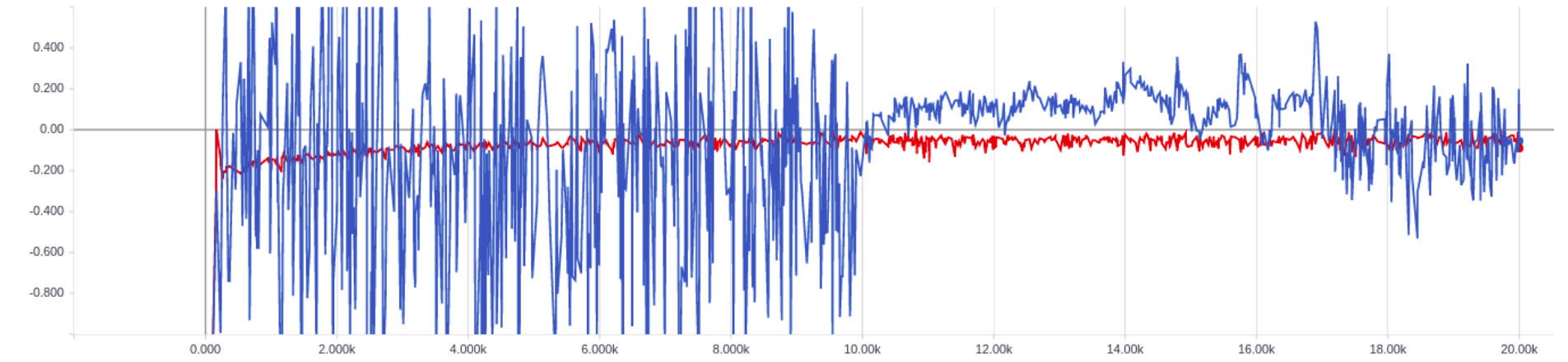}
\end{center}
\caption{Reproduction results of the original paper's Fig. 4. Evolution of the WGAN critic’s negative loss (without the regularization term) for $\lambda = 5$. Blue line: For the GP-penalty. Red line: For the LP-penalty.}
\end{figure}

Fig. 2 and Fig. 6 correspond to the reproduction results of this section. Again, we can see that the discussions in the experiment have been properly reproduced: WGAN-LP is robust to the selection of $\lambda$ than WGAN-GP, and critic function is learned much more stable.

\subsection{Estimating the Wasserstein distance}

\begin{figure}[h]
\label{fig3}
\begin{center}
\includegraphics[width=350px]{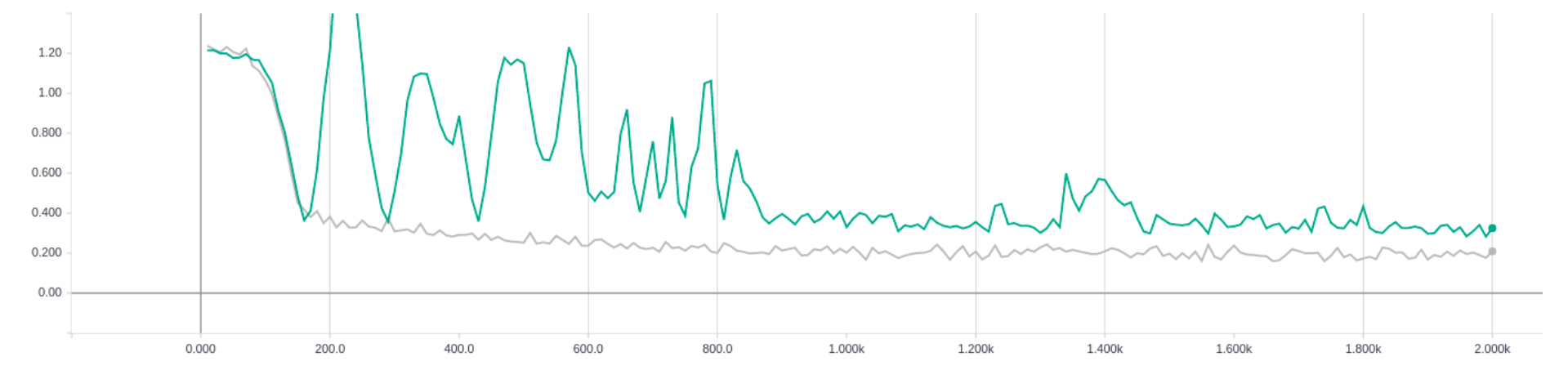}
\end{center}
\caption{Reproduction results of the original paper's Fig. 5. Evolution of the approximated Wasserstein-1 distance during training of WGANs on $\lambda = 5$. Mint line: For the GP-penalty. Gray line: For the LP-penalty}
\end{figure}

Fig. 2 and Fig. 6 correspond to the reproduction results of this section. At $\lambda = 5$, WGAN-LP model can generate a distribution with a smaller EMD faster than WGAN-GP model. On the other hand, WGAN-GP is not robust to the choice of lambda. Also, as discussed in the original paper, the WGAN-LP model shows stable EMD even at $\lambda = 100$.

\subsection{Different sampling methods}

Fig. 8 and Fig. 9 correspond to the reproduction results of this section. Slightly different from the discussion of the paper, we could not observe the difference in learning stability for the sampling method of the WGAN-GP model in a single run. However, in the top of original paper's Fig. 11, we can see that there is the case of remarkable fluctuation, despite showing overall stable trends. On the other hand, the middle of original paper's Fig. 11 exhibits overall unstable trends, but there are several extremely stable cases. Thus, failure to observe the stability difference according to sampling method in WGAN-GP is likely to be a sort of sampling error in our experiment.

\section{Conclusions}
We covered to reproduce the target paper on regularization for Wasserstein distance and showed all experiments are well reproducible. We wrote all the source code to reproduce the target paper. There is no existing source code for the target paper, so we made it based on various repositories related to WGAN. It took 3 days to write all the source code and reproduce the experiment. As mentioned above, we did not perform any further extended experiments in the review process. We have confirmed that the experimental results are reproducible in a verifiable and easy way. However, in some parts, it has been difficult to implement and reproduce.(e.g. 'How to implement 2-Wasserstein based regularization term in \emph{Tensorflow}') Thus, we prefer that authors who work on machine learning always publish their code. (for the target paper as well as for other papers)

We have confirmed what the target papers claimed: First, WGAN-LP has more stable learning and faster convergence property than WGAN-GP. Second, WGAN-LP is much more robust to regularization fraction $\lambda$ than WGAN-GP. Finding and determining the appropriate hyperparameter is an important but cumbersome, on study of machine learning. Therefore, presenting a robust model to the selection of hyperparameters can be a sufficient contribution to other researchers and the field itself. We are able to accept that the target paper contributes to this part in a reproducible way.

\bibliography{iclr2017_conference}
\bibliographystyle{iclr2017_conference}

\newpage
\section*{Appendix}

\begin{figure}[ht]
\label{fig4}
\begin{center}
\includegraphics[width=350px]{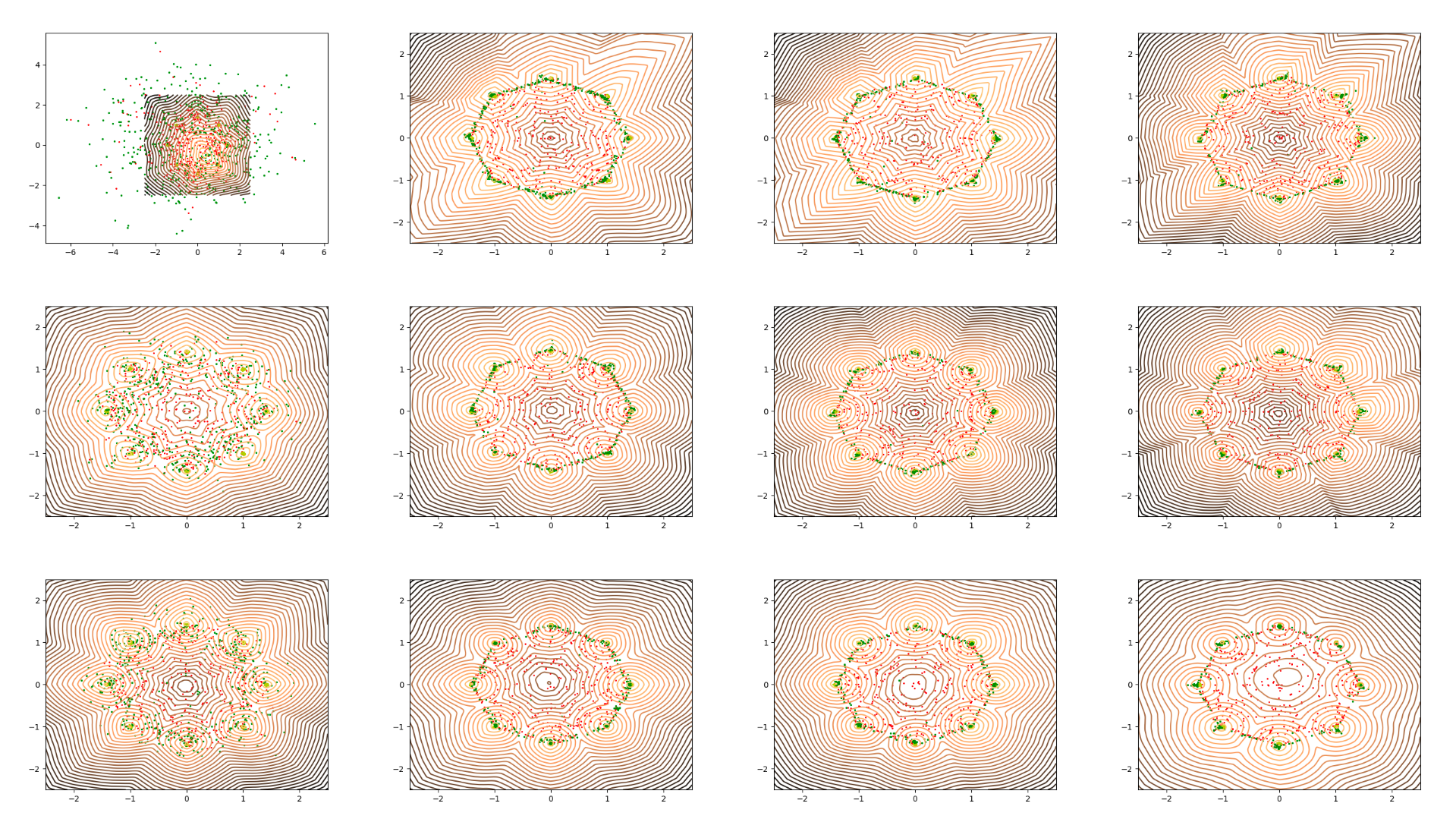}
\end{center}
\caption{Reproduction results of the original paper's Fig. 7. Level sets of the critic f of WGANs during training on \emph{8Gaussians} dataset. The same representation with Fig. 1 is used. Top: GP-penalty with $\lambda = 10$. Middle: GP-penalty with $\lambda = 1$. Bottom: LP-penalty with $\lambda = 10$.}
\end{figure}

\begin{figure}[ht]
\label{fig5}
\begin{center}
\includegraphics[width=350px]{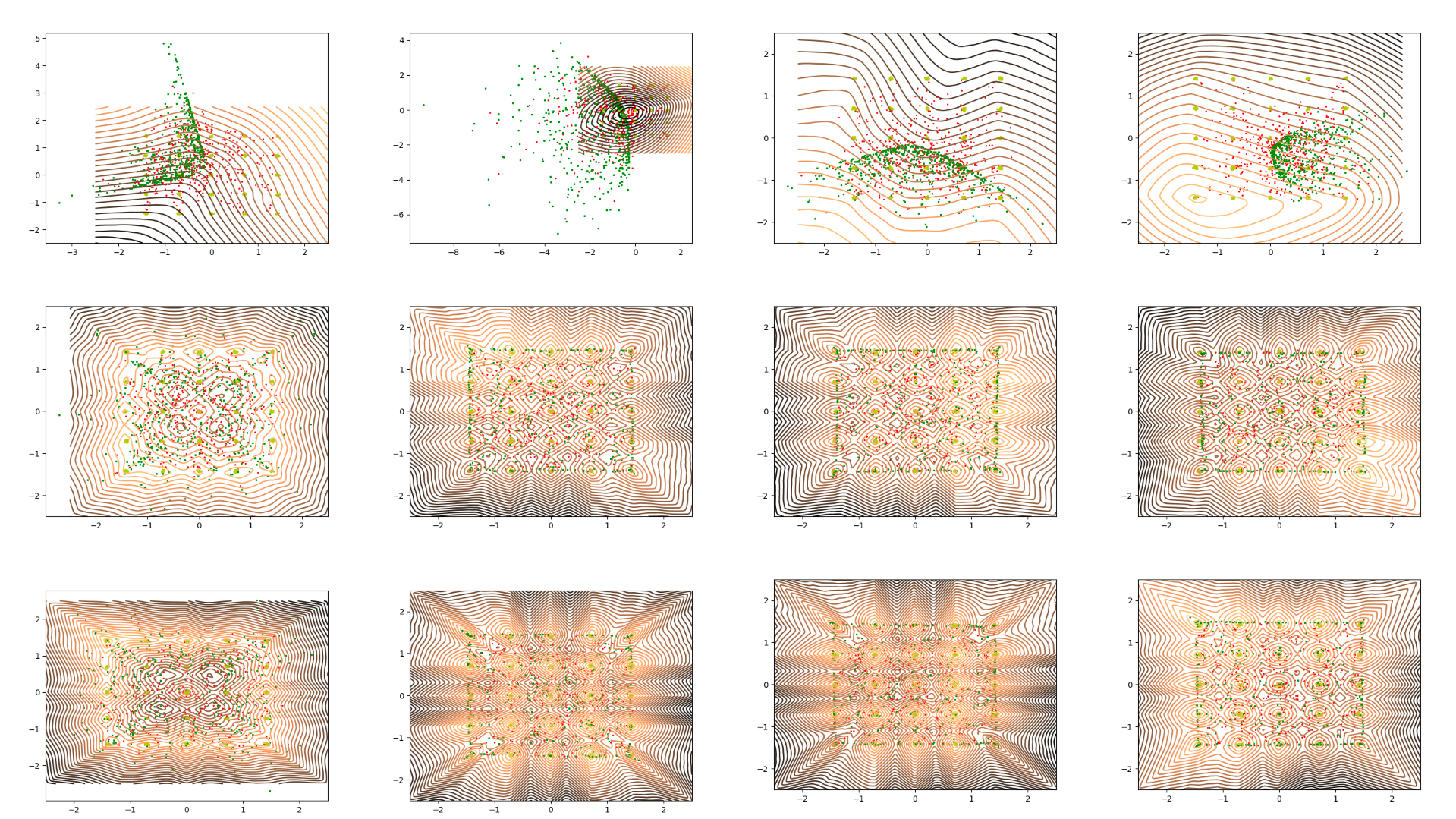}
\end{center}
\caption{Reproduction results of the original paper's Fig. 8. Level sets of the critic f of WGANs during training on \emph{25Gaussians} dataset. The same representation with Fig. 1 is used. Top: GP-penalty with $\lambda = 10$. Middle: GP-penalty with $\lambda = 1$. Bottom: LP-penalty with $\lambda = 10$.}
\end{figure}

\begin{figure}[ht]
\label{fig6}
\begin{center}
\includegraphics[width=350px]{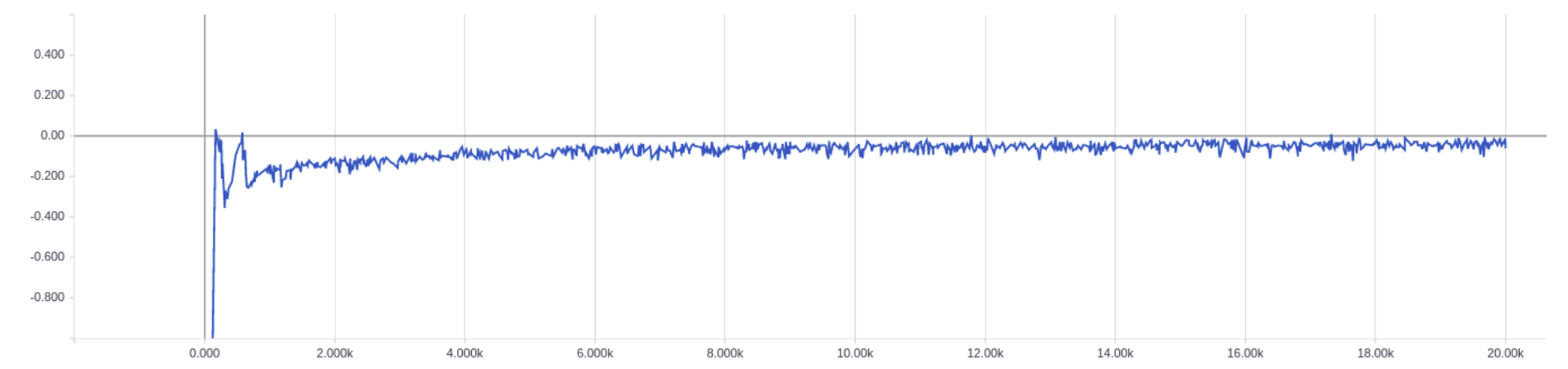}
\end{center}
\caption{Reproduction results of the original paper's Fig. 9. Evolution of the WGAN-GP critics loss without the regularization term on $\lambda = 1$.}
\end{figure}

\begin{figure}[ht]
\label{fig7}
\begin{center}
\includegraphics[width=350px]{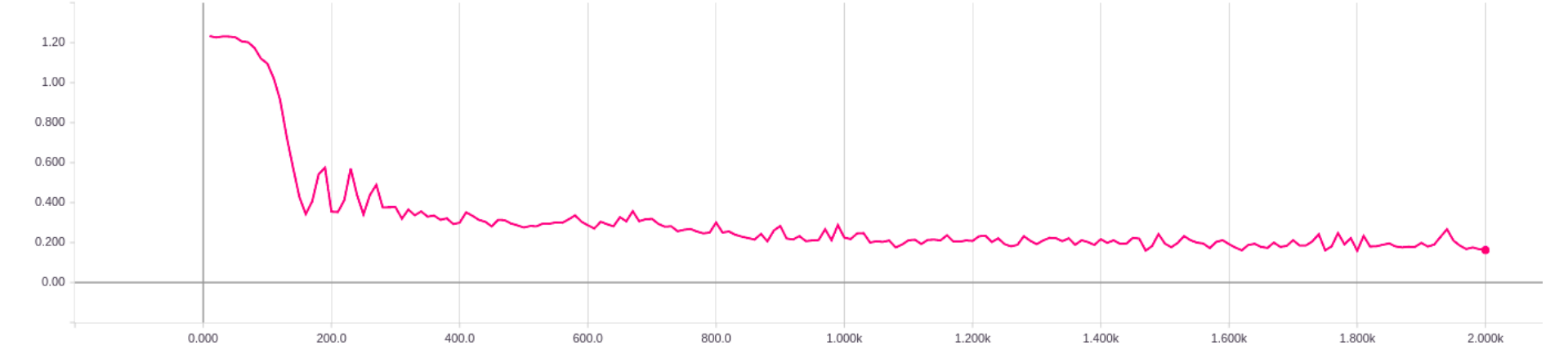}
\end{center}
\caption{Reproduction results of the original paper's Fig. 10. Evolution of the approximated EM distance during training WGAN-GPs on $\lambda = 1$.}
\end{figure}

\begin{figure}[ht]
\label{fig8}
\begin{center}
\includegraphics[width=350px]{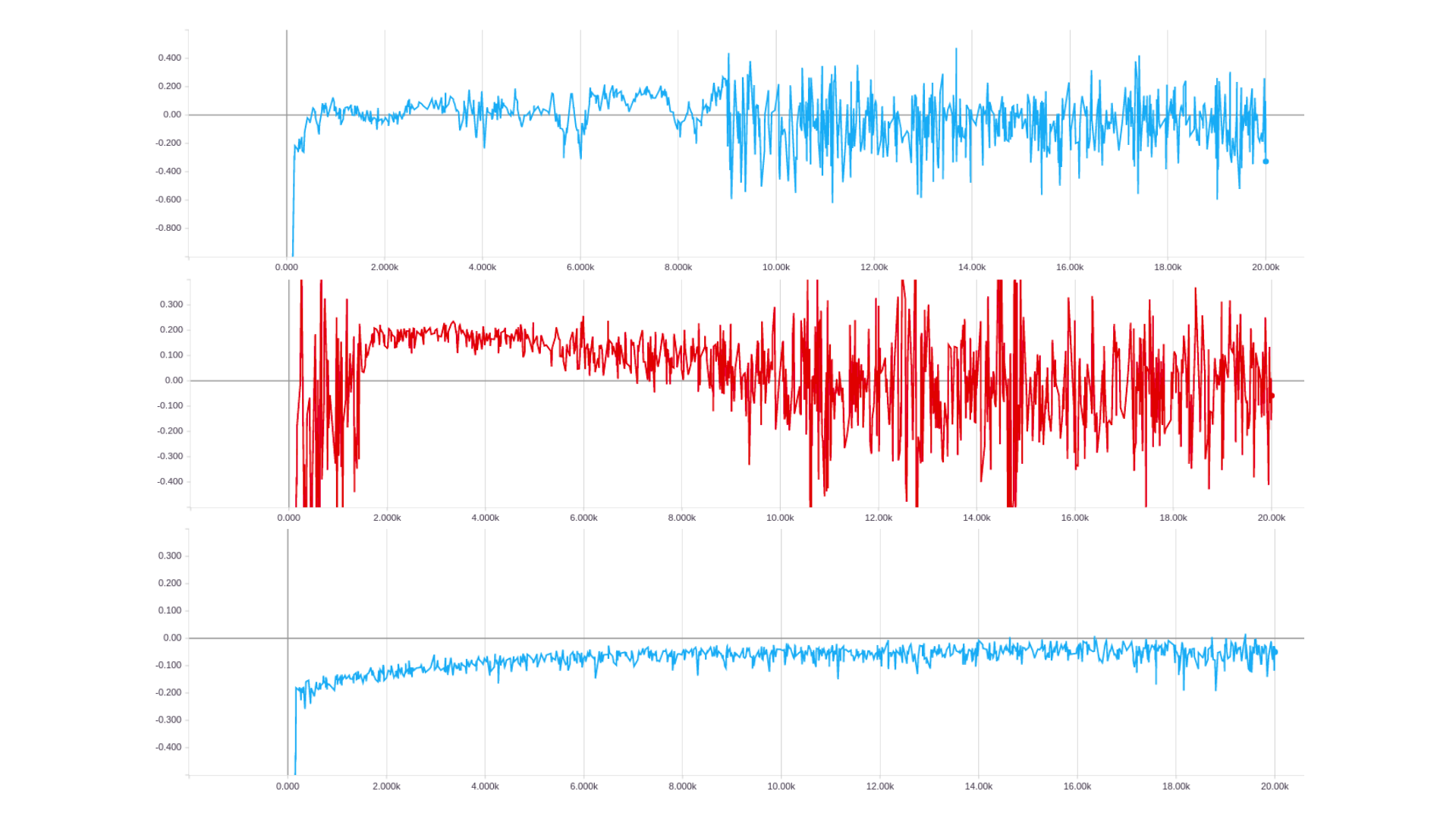}
\end{center}
\caption{Reproduction results of the original paper's Fig. 11. Evolution of the WGAN critic’s negative loss with local sampling (without the regularization term). Top: GP-penalty when generating samples by perturbing training samples only. Middle: For GP-penalty, perturbing training and generated samples. Bottom: LP-penalty, perturbing training and generated samples.}
\end{figure}

\begin{figure}[ht]
\label{fig9}
\begin{center}
\includegraphics[width=350px]{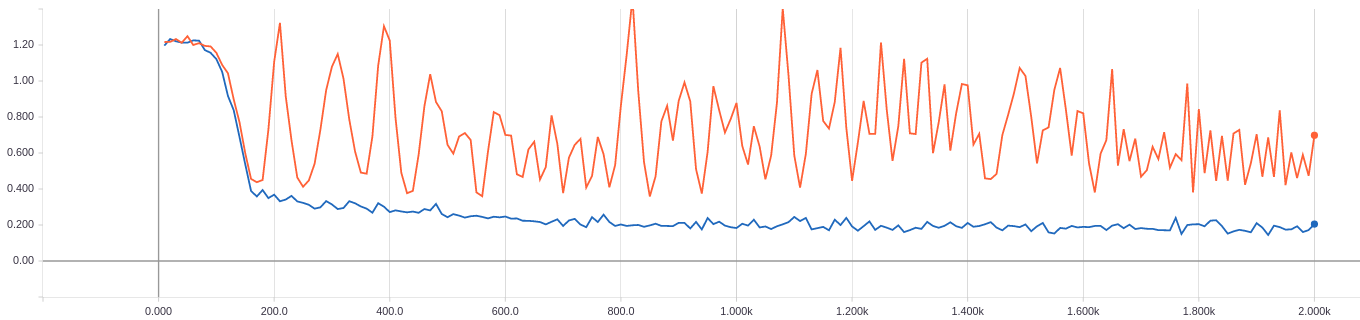}
\end{center}
\caption{Reproduction results of the original paper's Fig. 12. Evolution of the approximated EM distance during training of WGANs with local perturbation with $\lambda = 5$. Orange line: For the GP-penalty. Blue line: For the LP-penalty}
\end{figure}

\end{document}